\documentclass[journal]{IEEEtran}

\usepackage[colorlinks,urlcolor=blue,linkcolor=blue,citecolor=blue]{hyperref}

\usepackage{subfigure}
\usepackage{bigstrut}
\usepackage{graphicx}
\usepackage[table]{xcolor}

\usepackage{epstopdf}
\usepackage{url}
\usepackage{algorithm,algorithmic,cite,amsmath,amssymb,epsfig,enumerate,amsfonts,url,multirow,array,tabularx,booktabs,stfloats,flushend,color}
\usepackage{amsthm}
\usepackage{subfigure,balance}
\usepackage[T1]{fontenc}
\usepackage{booktabs}

\renewcommand\arraystretch{1.3}
\renewcommand{\algorithmicrequire}{\textbf{Input:}}

\def\x{{\mathbf x}}
\def\w{{\mathbf w}}
\def\y{{\mathbf y}}
\def\s{{\mathbf s}}
\def\u{{\mathbf u}}

\def\O{{\mathcal O}}

\def\bbx{{\bar {\mathbf{x}}}}

\def\bby{{\bar {\mathbf{y}}}}

\def\bbu{{\bar {\mathbf{u}}}}
\usepackage{amsmath,amssymb}
\newtheorem{thm}{Theorem}

\newtheorem{Cor}{Corollary}
\newtheorem{cor}[Cor]{Corollary}

\begin{document}

\title{fMBN-E: Efficient Unsupervised Network Structure Ensemble and Selection for Clustering}

\author{Xiao-Lei~Zhang
\thanks{Xiao-Lei Zhang is with the Research \& Development Institute of Northwestern Polytechnical University in Shenzhen, Shenzhen, China, and with the School of Marine Science and Technology,  Northwestern Polytechnical University, Xi'an, China. (e-mail: xiaolei.zhang@nwpu.edu.cn).}
}


\maketitle

\begin{abstract}
It is known that unsupervised nonlinear dimensionality reduction and clustering is sensitive to the selection of hyperparameters, particularly for deep learning based methods, which hinders its practical use. How to select a proper network structure that may be dramatically different in different applications is a hard issue for deep models, given little prior knowledge of data. In this paper, we aim to automatically determine the optimal network structure of a deep model, named multilayer bootstrap networks (MBN), via simple ensemble learning and selection techniques. Specifically, we first propose an MBN ensemble (MBN-E) algorithm which concatenates the sparse outputs of a set of MBN base models with different network structures into a new representation. Then, we take the new representation produced by MBN-E as a reference for selecting the optimal MBN base models. Moreover, we propose a fast version of MBN-E (fMBN-E), which is not only theoretically even faster than a single standard MBN but also does not increase the estimation error of MBN-E. Importantly, MBN-E and its ensemble selection techniques maintain the simple formulation of MBN that is based on one-nearest-neighbor learning. Empirically, comparing to a number of advanced deep clustering methods and as many as 20 representative unsupervised ensemble learning and selection methods, the proposed methods reach the state-of-the-art performance without manual hyperparameter tuning. fMBN-E is empirically even hundreds of times faster than MBN-E without suffering performance degradation. The applications to image segmentation and graph data mining further demonstrate the advantage of the proposed methods.
\end{abstract}



\setlength{\arraycolsep}{0.2em}
\section{Introduction}

\IEEEPARstart{U}{nsupervised} learning and clustering is a fundamental task of machine learning. It finds wide applications in data mining, text analysis, etc. Early works, like principal component analysis (PCA) and k-means clustering, conduct clustering in the original data space. Because the data in the original space is usually linearly-inseparable and noisy, later on, research turned to projecting data in the original space into a probability space where the data is supposed to be uniformly distributed and linearly separable, such as kernel methods, probabilistic models, and manifold and subspace learning \cite{fu2022latent}. However, a proper probability space is usually found by tuning parameters manually, e.g. kernel widths \cite{ng2001spectral} or regularization parameters, which is a long term headache problem. Although some work has tried to find the optimal parameters automatically, e.g. \cite{soares2004meta}, the learned representation, which is produced from a single layer nonlinear transform, is not abstract enough to describe the semantic classes of data.


 To learn highly abstract representations, deep neural network based data clustering has received much attention recently. The first work \cite{hinton2006reducing} extracts abstract representations from the bottleneck layer of a deep belief network. To make the deep representations suitable for clustering, some work adds additional terms, such as constraints \cite{huang2014deep}, clustering-like loss functions and models \cite{yang2017towards}, or novel network structures \cite{ji2017deep}, to the network training; while some work learns deep representations and refines cluster assignments iteratively \cite{xie2016unsupervised}. Recently, a new kind of deep learning based clustering, named self-supervised clustering optimizes cleverly designed objective functions of some pretext tasks, such as image completion, image colorization, or clustering, in which supervised pseudo labels are automatically obtained from the input data without manual annotations. It can be generally categorized into predictive self-supervised clustering  \cite{chang2018deep,wang2021progressive,wang2022local}, generative self-supervised clustering \cite{jiang2016variational,xia2021adversarial}, and contrastive self-supervised clustering  \cite{ji2019invariant,dang2021nearest}, respectively. See \cite{liu2021self} for an recent overview.
  Although the methods achieve superior performance over conventional clustering methods, many of them apply handcrafted priors to the benchmark data case by case, such as strong prior knowledge of data, data augmentation with clear intrinsic data structures, or hyperparameter tuning with the ground-truth labels. {If prior knowledge is insufficient, then some methods have to make a compromise with default hyperparameter settings, e.g. \cite{zhang2018multilayer}, which may degrade performance apparently.}



\begin{figure*}[t]
\centering
\resizebox{13cm}{!}{\includegraphics*{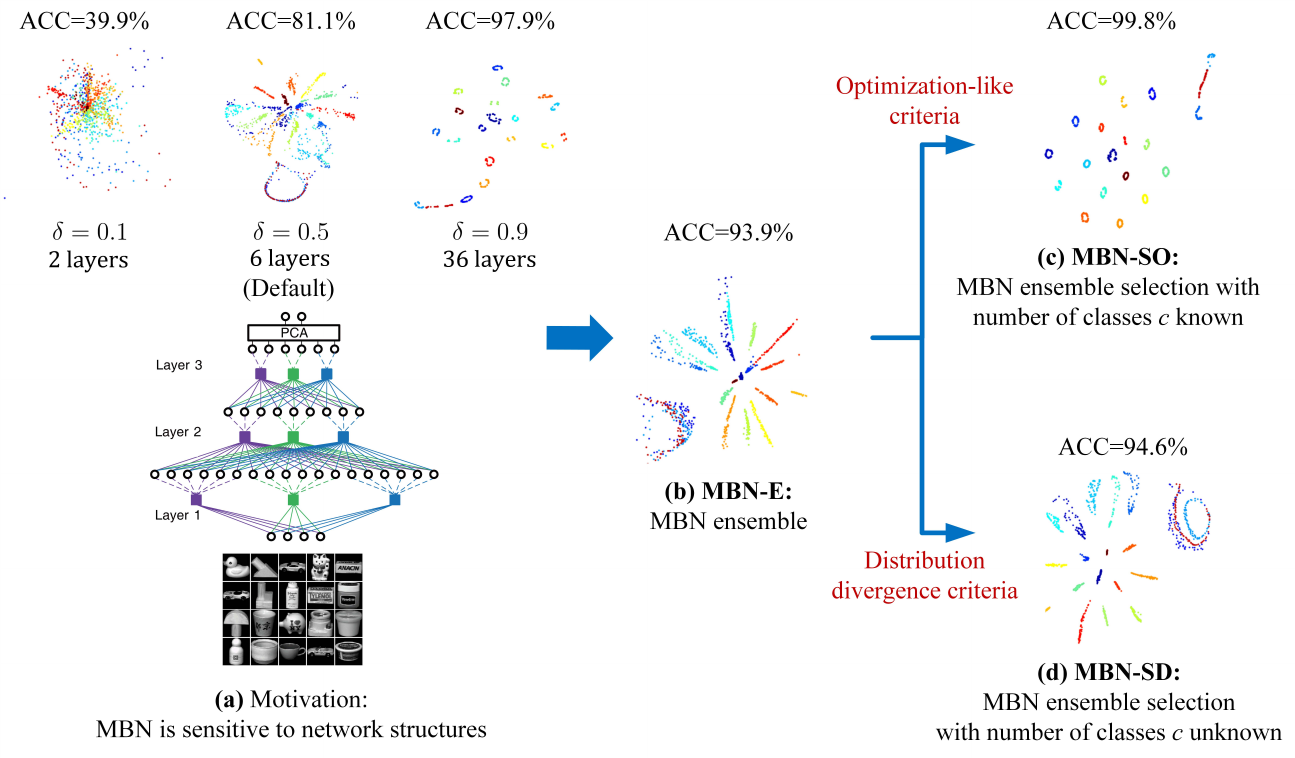}}
\caption{{On the network structure selection problem of MBN.} Each square of MBN in figure (a) represents a base clustering, while the black circles connected to the square represent the input/output of the base clustering. The hyperparameter ``$\delta$'' controls the network structure of MBN. The words in red color are two ensemble selection criteria for MBN-SO and MBN-SD respectively. The word ``ACC'' is short for clustering accuracy. The demo data is the COIL20 dataset \cite{nene1996columbia}.}
 \label{fig:1}
\end{figure*}

As we know, a long term goal of unsupervised learning and clustering is to design algorithms that are tuning-free and with little human labor, like k-means clustering. {However, from the above literature review, it seems that this topic is far from explored yet. Because the topic is a rather broad research area, this paper focuses on the network structure selection problem of a special deep model, named \textit{multilayer bootstrap network} (MBN) \cite{zhang2018multilayer}, given little prior knowledge of data. It seems a difficult problem, since that the network structure of MBN, which is controlled by hyperparameters, is strongly related to the unknown intrinsic property of the input data. See Section \ref{preliminary} for the details of the problem.}

We address the network structure selection problem of MBN by unsupervised ensemble learning and ensemble selection. See Section \ref{related} for an overview of the state-of-the-art works on unsupervised ensemble learning and selection. Although many ensemble selection methods may be applied successfully, we aim to exploit a simple and efficient way under the rule of Occam's Razor. We find empirically that, when applied to MBN, even a very simple ensemble selection method is able to achieve comparable top performance with the advanced ones, where 20 unsupervised ensemble learning and selection methods are used for comparison in Appendix D of the Supplement Material. Eventually, this finding derives a simple and efficient tuning-free unsupervised deep learning algorithm for practical use.

To summarize, as shown in Fig. \ref{fig:1}, the contribution of this paper is listed as follows:
\begin{itemize}
\item We theoretically prove that increasing the depth of MBN does not always improve the performance, which induces the network structure selection problem of MBN.

  \item To address the aforementioned problem, we propose a simple MBN ensemble (MBN-E) algorithm. It groups the sparse outputs of a number of MBN base models with different network structures into a new representation.

    \item To reduce the high computational complexity problem of MBN-E, we propose the fast MBN-E (fMBN-E) by a simple modification of MBN-E. 
        It accelerates MBN-E by over hundreds of times both theoretically and empirically. We have proved that the acceleration does not degrade the performance. 

  \item To further improve the performance of MBN-E, we propose (i) the MBN ensemble selection with optimization-like criteria (MBN-SO) for the case when the number of classes is known, and (ii) the MBN ensemble selection with distribution divergence criteria (MBN-SD) when the number of classes is unknown. Both of them select a number of highly-effective MBN base models from MBN-E to group into a new MBN-E. The difference between them lies in the selection criteria of the base models.

  \item We have run experiments on a number of benchmark datasets where the optimal network structure of MBN appears in fundamentally different ranges. 
 Experimental results show that MBN-E significantly outperforms the MBN with the default setting and approaches to the MBN with the optimal setting. fMBN-E achieves similar performance with MBN-E, and is over dozens of times faster than MBN-E. MBN-SO and MBN-SD further improves the performance of MBN-E.

 \item Because the proposed algorithms intend to solve the difficulty of real-world applications of MBN, we further applied the proposed methods to image segmentation and graph data mining. Experimental results verified the effectiveness of the proposed methods.

%
\end{itemize}

The rest of the paper is organized as follows. In Section \ref{related}, we present related work. In Section \ref{preliminary}, we review MBN. In Section \ref{sec:anal}, we analyze the structure selection problem of MBN both theoretically and empirically. In Sections \ref{mbn-e} and \ref{mbn-s}, we present MBN-E, fMBN-E, MBN-SO, and MBN-SD, respectively. In Section \ref{experiment}, we present an extensive experiment. In Section \ref{sec:appl}, we apply the proposed methods to image segmentation and graph data mining. Finally, in Section \ref{conclusion}, we conclude the paper.

 \section{Related work}\label{related}
 The proposed MBN-E essentially is rooted in clustering ensemble. The proposed MBN-SO and MBN-SD are essentially rooted in ensemble selection and reweighting. The selection criteria of the base models of MBN-SD, which measures the divergence between data distributions, are rooted in unsupervised domain adaptation. We present the three aspects as follows.

\subsection{Clustering ensemble}

Ensemble learning, such as {\it{bagging}}, {\it{boosting}}, and their variations, has demonstrated its effectiveness on many learning problems \cite{dietterich2000ensemble}.
Unsupervised ensemble learning inherits the fundamental theories and methods of classifier ensemble. The mostly studied unsupervised ensemble learning is \textit{clustering ensemble}. It
aims to combine multiple \textit{base clusterings} with a so-called \textit{meta-clustering function}, a.k.a \textit{consensus function}, for enhancing the stability and accuracy of the base clusterings \cite{strehl2003cluster,vega2011survey}. Meta-clustering functions can be categorized generally to two classes \cite{vega2011survey}. The first class analyzes the co-occurrence of objects: how many times an object belongs to one cluster or how many times two objects belong to the same cluster. The second class, called the median partition, pursues the maximal similarity with all partitions in the ensemble {\cite{li2007solving,nguyen2007consensus,yu2020clustering}}. Recently, some unsupervised deep ensemble learning methods have been proposed \cite{liu2016infinite,koohzadi2020unsupervised,hu2022representation}. For example, \cite{liu2016infinite} takes deep neural networks act like a meta-clustering function. \cite{koohzadi2020unsupervised} decomposes each layer of a deep neural network into an ensemble of encoders or decoders and mask operations.

To our knowledge, unsupervised deep ensemble learning is not prevalent, due to maybe that neural networks need supervised signals to maximize their discriminant ability. See \cite{vega2011survey,ganaie2021ensemble} for the reviews of clustering ensemble.




\subsection{Clustering ensemble reweighting and selection}

Because not all base clusterings contribute equivalently to a cluster ensemble, it is needed to conduct ensemble reweighting and selection, which mainly focuses on three respects: (i) different types of weights, (ii) algorithms for calculating the weights, and (iii) cluster validation criteria for measuring the diversity and quality of the base models.

The most common type of weights is to assign a weight to each base clustering according to its quality or/and diversity in the ensemble, e.g. \cite{zhou2006clusterer}. A special case of this type is to constrain the weights of some weak base clusterings to zero, named \textit{clustering selection} \cite{fern2008cluster,azimi2009adaptive}. However, weak base clusterings may also contain some high quality clusters, and vise versa. With this perspective, many reweighting strategies at levels of clusters \cite{yang2010temporal,huang2017locally}, data structures \cite{yu2016distribution}, and data points \cite{li2019clustering} were proposed.

The algorithms for calculating the weights can be categorized into two types \cite{zhang2019weighted}. The first type calculates weights by measuring the similarity between the predicted labels of the clustering ensemble and its base clusterings \cite{zhou2006clusterer,fern2008cluster}. The second type treats the weights as variables of consensus functions which are obtained by advanced optimization algorithms, e.g. \cite{li2008weighted}.

The criteria for measuring the diversity and quality of the base models can be categorized into two classes. The first class of measurements calculates the normalized mutual information \cite{zhou2006clusterer,fern2008cluster}, adjusted rand index \cite{jia2011bagging}, clustering accuracies \cite{hong2009resampling}, and their variants \cite{duarte2006weighted} or aggregations \cite{huang2021toward} between the sets of the predicted labels. The second class of validation criteria is based on data distributions \cite{vendramin2010relative,halkidi2001clustering}. They usually calculate some kinds of statistics of data \cite{naldi2013cluster,yu2016distribution}. Some systematical studies on cluster validation indices \cite{vendramin2010relative,halkidi2001clustering} have been carried out as well.


To summarize, when the number of classes is given, we evaluate the quality of the base models by \textit{optimization-like criteria} \cite{vendramin2010relative}, for MBN-SO. When the number of classes is not given, we propose to evaluate the quality of the base models by so-called \textit{distribution divergence criteria} for MBN-SD, which measure the learned representations of data directly without predicted labels.

\subsection{Unsupervised domain adaptation}

Domain adaptation is the ability of applying an algorithm trained in one or more ``source domains'' to a different but related ``target domain''. Unsupervised domain adaptation is a subtask of domain adaptation where the target domain does not have labels. The algorithms can be categorized into three branches \cite{kouw2019review}, which are sample-based, feature-based, and inference-based approaches. No matter how the approaches vary, the distribution divergence measurement between the source domains and the target domain always lies in the core of unsupervised domain adaptation. The most popular measurement is maximum mean discrepancy (MMD) \cite{borgwardt2006integrating}. Other measurements include Kullback-Leibler divergence, total variation distance, second-order (covariance) statistics, and Hellinger distance. Although the distribution divergence measurement has been extensively studied in unsupervised domain adaptation, it seems far from explored in unsupervised ensemble selection. 

In this paper, we name this kind of measurements as distribution divergence criteria, and apply them to MBN-SD. Because MMD performs generally well among the measurements and is applicable to all data types, from high-dimensional vectors to strings and graphs, we focus on using MMD.

 \section{Preliminaries}\label{preliminary}

 This section presents MBN and its theoretical foundation briefly. See Appendices A and B of the Supplementary Material for the summary of important notations and detailed description of MBN as well as its geometric and theoretical foundations.

\subsection{Multilayer bootstrap networks}

This paper takes MBN \cite{zhang2018multilayer} as a research object. It is a simple deep model. As shown in Fig. \ref{fig:1}a, suppose we are to build an $M$-layer MBN from bottom-up, it can be described as follows:
\begin{itemize}
  \item Step 1, for each layer, MBN trains $V$ mutually-independent $k$-centroids base clusterings, where the parameter $k$ of all clusterings at the same layer is the same. For each base clustering, it takes the following three operators successively to generate a new representation of data:
      \begin{itemize}
      \item \textbf{Random selection of features:} It first randomly selects some features of the input data, which yields a new representation of the data.
      \item \textbf{Random sampling of data:} It randomly samples $k$ data points from the data with the new representation as the $k$ centroids.
      \item \textbf{One nearest neighbor optimization:} It assigns each input data to one of the $k$ clusters, and outputs a $k$-dimensional one-hot code, indicating which cluster the input data belongs to.
      \end{itemize}
      The one-hot representations from all base clusterings are concatenated as the input of the upper layer.
  \item Step 2, MBN stacks the cluster ensemble described in Step 1 for $M$ times. The parameter $k$ at two adjacent layers have the following connection:
  \begin{equation}\label{eq:delta}
    k_{m} = \delta k_{m-1}
  \end{equation}
  where $k_m$ and $k_{m-1}$ are the parameter $k$ at the $m$-th and $(m-1)$-th adjacent layers respectively, and $\delta\in(0,1)$ is a hyperparameter controlling the network structure of MBN. Because $\delta\in(0,1)$, we must have
 \begin{equation}\label{eq:k}
   k_1>k_2>\ldots>k_m>\ldots >k_o
 \end{equation}
where $k_o$ is the parameter $k$ at the top layer. Note that, the total number of layers of MBN is usually determined automatically by $k_1$, $k_o$, and $\delta$.
\end{itemize}

\subsection{Estimation error of a single layer of MBN}

\cite{zhang2018multilayer} analyzed the estimation error of a single layer of MBN, which explains the empirical success of MBN. We summarize the analysis here.

Given an input $\x$ of MBN at a layer, it is easy to image that each $k$-centroids clustering contributes a nearest neighbor $\w_v$ to $\x$, $\forall v=1,\ldots,V$, then, the new location of $\x$ in the input data space, denoted as $\hat{\x}$, is given by the $V$ nearest neighbors as:
\begin{eqnarray}\label{eq:soafaj}
\hat{\x} = \frac{1}{V}\sum_{v=1}^V \w_v
\end{eqnarray}
If $\hat{\x}$ is an effective estimation of $\x$, then the \textit{locally linear assumption} between $\{\w_v\}_{v=1}^V$ and $\x$ must hold; otherwise, $\hat{\x}$ is not an accurate estimation.

Under the locally linear assumption, the estimation error $\mathbb{E}(x-\hat{x})$ can be decomposed into the following form using the famous \textit{bias-variance decomposition of expectation risk} \cite{hastie2009unsupervised}:
   \begin{eqnarray}\label{eq:bv}
\mathbb{E}((\x-\hat{\x})^2)& = &(\x-\mathbb{E}(\hat{\x}))^2 + \mathbb{E}\left((\x-\mathbb{E}(\hat{\x}))^2\right)\nonumber\\
&=&\mathrm{Bias}^2(\hat{\x}) + \mathrm{Var}(\hat{\x})
 \end{eqnarray}
 Given \eqref{eq:bv}, we can derive the following theorem for the estimation error of a single layer of MBN:
\begin{thm}\label{thm:4}
  {
  The estimation error of a single layer of MBN $\mathbb{E}_{\mathrm{ensemble}}$ and the estimation error of a single $k$-centroids clustering $\mathbb{E}_{\mathrm{single}}$ in the layer have the following relationship:
 \setlength{\arraycolsep}{0.2em}
   \begin{eqnarray}\label{eq:important}
\mathbb{E}_{\mathrm{ensemble}} = \left(\frac{1}{V}+\left(1-\frac{1}{V}\right)\rho\right)\mathbb{E}_{\mathrm{single}}
 \end{eqnarray}
where $\rho$ is the pairwise positive correlation coefficient between the $k$-centroids clusterings, $0\leq \rho \leq 1$ \cite{zhang2018multilayer}.}
\end{thm}

\section{Analysis of the network structure problem of MBN}\label{sec:anal}

It is expected that adding more layers to a deep network could improve the representation learning ability of the network. However, this is not always the case empirically, so as to MBN.

In this section, we first give an empirical demo on how different network structures affect the performance in Section \ref{subsec:empirical}, and then derive the estimation error of the entire MBN in Section \ref{subsec:theoretical} by extending Theorem \ref{thm:4} to the multilayer scenario, which explains the empirical phenomenon theoretically and motivates the novel algorithms of this paper.

\subsection{{Empirical justification}}\label{subsec:empirical}

A core problem of MBN is that its effectiveness is strongly related to the network structure which is controlled by parameter $\delta$. Given parameters $k_1$ and $k_o$ in \eqref{eq:k} fixed, how fast $k$ drops from $k_1$ to $k_o$ layer by layer according to \eqref{eq:k}, which is determined by $\delta$, should match the nonlinearity and noise level of data. When $\delta$ approaches to 0, MBN builds a shallow network with a single nonlinear layer, which is suitable for linearly separable data. When $\delta$ is enlarged towards 1, MBN becomes deeper and deeper, which is suitable for highly nonlinear and non-Gaussian data. If the above regularity is violated, the performance of MBN may drop sharply.

\begin{figure}[t]
\centering
\resizebox{8.5cm}{!}{\includegraphics*{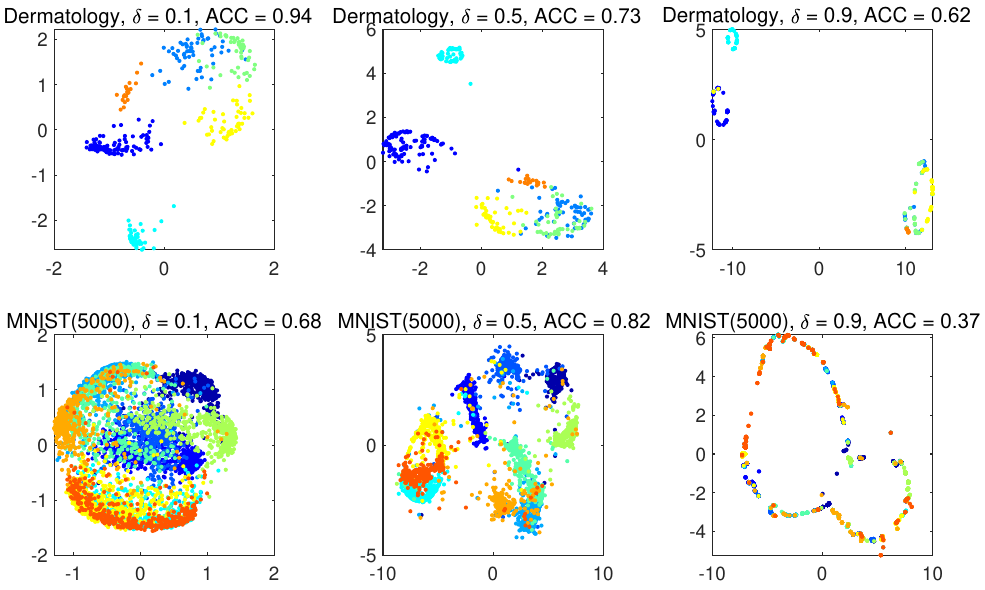}}
\caption{{Visualization of features produced by MBN with different $\delta$ on the Dermatology and MNIST(5000) datasets, where Dermatology is a dataset from UCI, and MNIST(5000) is a subset of MNIST dataset that consists of 5000 randomly selected data points.} }
 \label{fig:new}
\end{figure}

In Fig. \ref{fig:1}a, we can see that, increasing $\delta$ from 0.1 to 0.9 yields gradually improved performance on COIL20. The gap between the best performance and poorest performance is as high as 58\%. However, in Fig. \ref{fig:new}, we see that (i) the best performance of MBN on the Dermatology dataset appears at $\delta = 0.1$, and the performance degrades gradually along with the increase of $\delta$, which is contrary to the trend on COIL20; (ii) the best performance on MNIST(5000) appears at $\delta = 0.5$, which significantly outperforms the performance when $\delta = 0.1$ and $\delta = 0.9$. Moreover, as will be shown in Table \ref{table:data_set_info} and Fig. \ref{fig:scores} in the experiment, the best $\delta$ for different datasets appears at dramatically different ranges.

Because it is difficult to evaluate the properties of data in unsupervised learning, MBN has to make a compromise by setting $\delta = 0.5$. This may lead to far inferior performance from the optimal one, though $\delta = 0.5$ happens to be the best choice on some data like MNIST. In this paper, we aim to address this issue by detecting the optimal $\delta$ automatically.

\subsection{Theoretical explanation}\label{subsec:theoretical}

\begin{figure}[t]
\centering
\resizebox{8.5cm}{!}{\includegraphics*{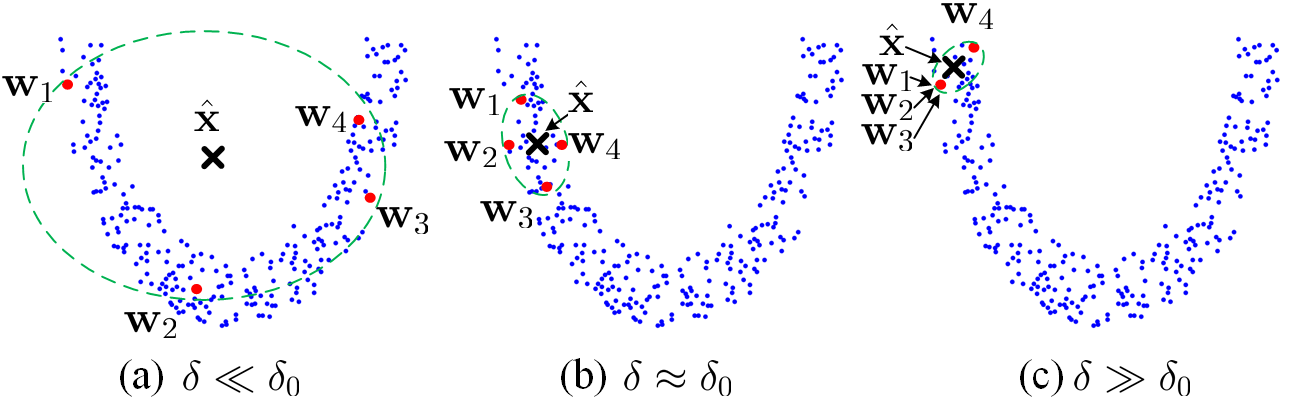}}
\caption{{Diagram of the density estimation process of MBN with different $\delta$. The notation $\delta_0$ denotes the optimal $\delta$. The black cross $\hat{\mathbf{x}}$ denotes the coordinate of the learned representation of an input data $\x$. The four red points, which are $\w_1$, $\w_2$, $\w_3$ and $\w_4$ respectively, are the nearest centroids of four $k$-centroids clusterings to an input data point $\x$. The blue dotted oval is the area of the locally linear assumption.} }
 \label{fig:fajo}
\end{figure}

 A fundamental element of MBN is the locally linear assumption defined in \eqref{eq:soafaj}. The correctness of the assumption is strongly related to the choice of $\delta$. Suppose the optimal performance of MBN appears at $\delta = \delta_0$. Then, a diagram in Fig. \ref{fig:fajo} explains the empirical phenomenon in Section \ref{subsec:empirical}.

When we set $\delta \ll \delta_0$, the locally linear assumption \eqref{eq:soafaj} may be violated, which makes MBN fail to learn correct representations. For example, in Fig. \ref{fig:fajo}a, given an input data point $\x$ that is sampled from the nonlinear data distribution, its representation $\hat{\mathbf{x}}$ learned by the nearest centroids $\w_1$, $\w_2$, $\w_3$, and $\w_4$ is even out of the data distribution, which is clearly wrong. This explains the empirical phenomenon that MBN does not reach the top performance on COIL 20 when $\delta\ll 0.9$,  and on MNIST(5000) when $\delta\ll 0.5$.

\begin{figure}[t]
\centering
\resizebox{5.5cm}{!}{\includegraphics*{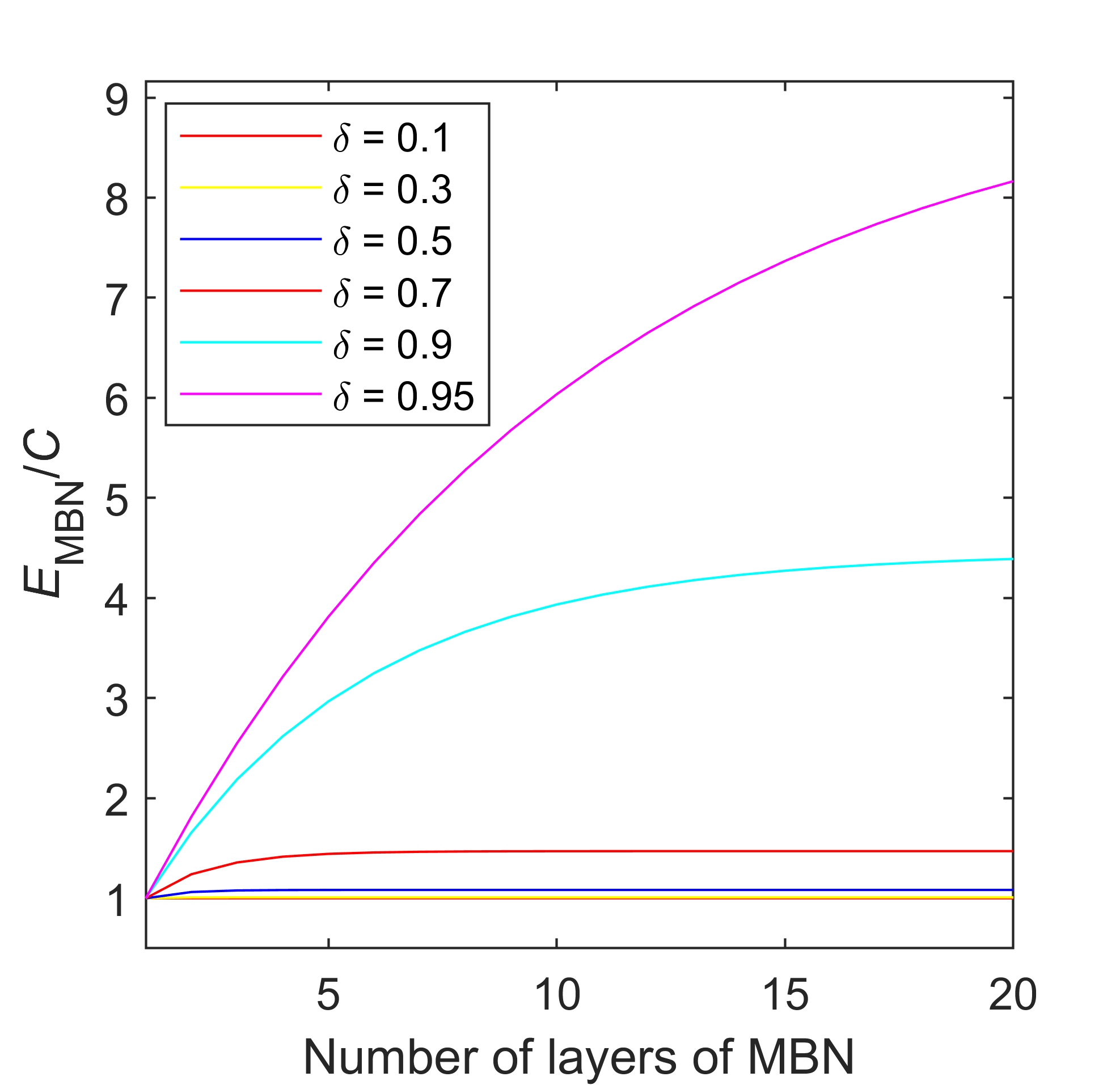}}
\caption{{Connection between the estimation error of MBN and $\delta$ when $\delta>\delta_0$, where $C = \left({a k_1}/{n}\right)^2\mathbb{E}_{(\mathrm{single,1})}$ is a constant.} }
 \label{fig:fajofa}
\end{figure}

To explain the failure of MBN at $\delta \gg \delta_0$, we first give the following theorem:
\begin{thm}\label{thm:MBN_error}
When $\delta >\delta_0$, the estimation error of MBN is:
      \begin{equation}\label{eq:error_accumulate}
\mathbb{E}_{\mathrm{MBN}} \geq \sum_{m=1}^M \left(\frac{1}{V}+\left(1-\frac{1}{V}\right)\left(\frac{a k_1}{n}\right)^2\delta^{2(m-1)}\right)\mathbb{E}_{(\mathrm{single,1})}
 \end{equation}
where $a\in(0,1]$ is the ratio of the number of randomly selected features over the number of all features in Step 1 of MBN, $\mathbb{E}_{\mathrm{single,1}}$ is the estimation error of a single $k$-centroids clustering at the bottom layer, and $M$ is the number of nonlinear layers of MBN.
\end{thm}
\begin{proof}
First of all, we should emphasize that, when $\delta <\delta_0$, the locally linear assumption for \eqref{eq:soafaj} does not hold, which makes Theorem \ref{thm:4} do not hold as well. Because the following proof is built on Theorem \ref{thm:4}, Theorem \ref{thm:MBN_error} is effective only when $\delta >\delta_0$.

Because the probability that any two $k$-centroids clusterings select the same element of the same input data point as one of their centroids is $(ak/n)^2$, then we can imagine easily that the correlation is
     \begin{eqnarray}\label{eq:rho}
\rho=(ak/n)^2
 \end{eqnarray}
 We denote the correlation at the $m$th layer as $\rho_m$. Substituting \eqref{eq:delta} into \eqref{eq:rho} derives
      \begin{eqnarray}\label{eq:connection}
\rho_{m} = (a k_{m-1}/n)^2\delta^2=\ldots=  (a k_{1}/n)^2\delta^{2(m-1)}
 \end{eqnarray}
We denote the estimation error of a single $k$-centroids clustering and an ensemble of clusterings at the $m$th layer as $\mathbb{E}_{({\mathrm{single}},m)}$ and $\mathbb{E}_{({\mathrm{ensemble}},m)}$ respectively. Because reducing $k$ makes $\mathbb{E}_{{\mathrm{single}}}$ enlarged, we may assume that $\mathbb{E}_{({\mathrm{single}},m)}$ is lower-bounded by $\mathbb{E}_{({\mathrm{single}},1)}$. Substituting \eqref{eq:connection} into \eqref{eq:important} derives:
      \begin{equation}\label{eq:error_accumulatex}
\mathbb{E}_{(\mathrm{ensemble},m)} \geq \left(\frac{1}{V}+\left(1-\frac{1}{V}\right)\left(\frac{a k_1}{n}\right)^2\delta^{2(m-1)}\right)\mathbb{E}_{(\mathrm{single,1})}
 \end{equation}
Because $\mathbb{E}_{\mathrm{MBN}}$ accumulates $\mathbb{E}_{(\mathrm{ensemble},m)}$ of all layers from bottom-up, we can derive the overall estimation error of MBN as \eqref{eq:error_accumulate}.
\end{proof}
We further derive the following corollary from Theorem \ref{thm:MBN_error}:
\begin{cor}\label{cor:faoo}
When $\delta>\delta_0$ and $V\rightarrow \infty$, the estimation error of MBN is:
 \begin{equation}
\mathbb{E}_{\mathrm{MBN}} \geq C \sum_{m=1}^M \delta^{2(m-1)}
 \end{equation}
 where $C = \left({a k_1}/{n}\right)^2\mathbb{E}_{(\mathrm{single,1})}$ is a constant.
\end{cor}
Corollary \ref{cor:faoo} can be visualized in Fig. \ref{fig:fajofa}. From the figure, we see that, when $\delta$ approaches to 1, $\mathbb{E}_{\mathrm{MBN}}$ is increased exponentially.

Fig. \ref{fig:fajo}c gives an example on how the large estimation error occurs when $\delta\gg\delta_0$. In this figure, we see that, because the four $k$-centroids clusterings have strong correlation, three out of four nearest centroids to $\x$, i.e. $\w_1$, $\w_2$, and $\w_3$, share the same location, which makes MBN difficult to learn a good representation. The above analysis explains the phenomenon why the performance of MBN on Dermatology and MNIST(5000) drops sharply when $\delta=0.9$.

As shown in Fig. \ref{fig:fajo}b, only when $\delta\approx \delta_0$, not only the locally linear assumption holds, but also the $k$-centroids clusterings have weak correlation, which makes MBN learn the best representation for $\x$. However, avoiding the sensitivity of MBN to $\delta$ is not straightforward, which motivates the proposed methods in the following of this paper.

\section{Multilayer bootstrap network ensemble}\label{mbn-e}

In this section, we first introduce MBN-E in Section \ref{mbn-e1}, then present an efficient algorithm for MBN-E, named fMBN-E, in Section \ref{fMBN-E}, and finally discuss why fMBN-E can accelerate MBN-E without degrading the estimation accuracy in Section \ref{discussion}.

\subsection{MBN-E}\label{mbn-e1}
Because MBN is sensitive to $\delta$, a straightforward thought is to integrate a number of MBN base models with different $\delta$ into MBN-E. We present MBN-E in Algorithm \ref{alg:mbn-e}.

In Algorithm \ref{alg:mbn-e}, we usually conduct PCA preprocessing to $\{\x_i \}_{i=1}^{n}$ before MBN-E, which not only reduces the computational complexity of the bottom layers of the MBN base models but also de-correlates the input features. After getting the output $\{\bby_{i}\}_{i=1}^n$, we sometimes need to reduce $\{\bby_{i}\}_{i=1}^n$ to a low-dimensional representation $\{\bbu_{i}\}_{i=1}^n$ in an Euclidian space by, e.g. PCA, for applications, since that $\{\bby_{i}\}_{i=1}^n$ is very high dimensional. Likewise, we denote the low-dimensional representation of the base models $\{\y_{z,i}\}_{i=1}^n$ as $\{\u_{z,i}\}_{i=1}^n$.

The computational complexity of MBN-E, which is $Z$ times higher than MBN, is too high to be intolerable in practice when $Z\gg1$:
\begin{thm}\label{thm:2}
  {The computational complexity of MBN-E approximates to $ Z(\O(\alpha kVn)+\O(kVn))$ empirically, where $\O(\alpha kVn)$ and $\O(kVn)$ are the complexity of a single MBN at the bottom layer and the other layers respectively, and $\alpha$ is a constant related to the sparse property of the input data.}
\end{thm}

 \begin{algorithm}[t]
    \caption{MBN-E.}
    \begin{algorithmic}[1]\label{alg:mbn-e}
\REQUIRE  A $h$-dimensional unlabeled dataset $\{\x_i \}_{i=1}^{n}$, parameter $k_o$, and number of MBN base models $Z$\\
\renewcommand{\algorithmicrequire}{\textbf{Initialization: }}
\ENSURE $\{\bar{\y}_i \}_{i=1}^{n}$
\FOR{$z=1,\ldots,Z$}
\STATE Randomly generate $\delta$ from the range $[0.05,0.95]$;
\STATE $\{\y_{z,i} \}_{i=1}^{n}\leftarrow \mathrm{MBN}(\{\x_i \}_{i=1}^{n}, k_o,\delta)$\\
\ENDFOR
\FOR{$i=1,\ldots,n$}
    \STATE $\bar{\y}_{i}\leftarrow [\y^T_{1,i},\y^T_{2,i},\ldots,\y^T_{Z,i}]^T$
\ENDFOR
\end{algorithmic}
\end{algorithm}

\begin{figure}[t]
\centering
\resizebox{7.8cm}{!}{\includegraphics*{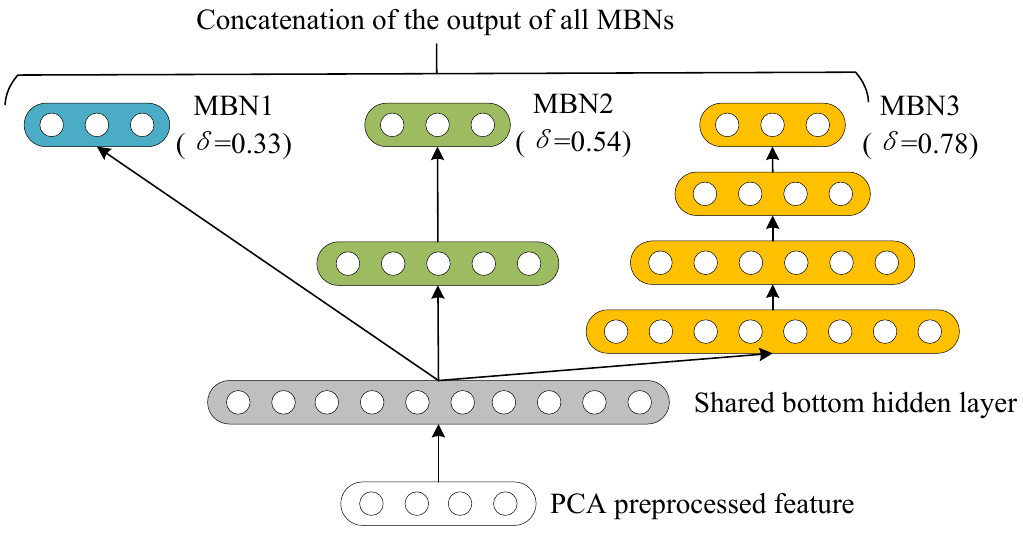}}
\caption{{Architecture of fMBN-E.} Different color represents different MBN base models with random $\delta$ values.}
 \label{fig:x}
\end{figure}

\subsection{fMBN-E}\label{fMBN-E}


 \begin{algorithm}[t]
    \caption{fMBN-E.}
    \begin{algorithmic}[1]\label{alg:fmbn-e}
\REQUIRE  A $h$-dimensional unlabeled dataset $\{\x_i \}_{i=1}^{n}$, parameter $k_o$, and number of MBN base models $Z$\\
\renewcommand{\algorithmicrequire}{\textbf{Initialization: }}
\REQUIRE{
$k_1= \lfloor n/2\rfloor$,
number of base clusterings per layer $V=400$
}
\ENSURE $\{\bar{\y}_i \}_{i=1}^{n}$\\
\STATE /* train a shared bottom layer */
\STATE $\{\y_{i} \}_{i=1}^{n}\leftarrow \mathrm{MBN}(\{\x_i \}_{i=1}^{n}, k_1-1,\delta=0)$\\
\STATE /* train an ensemble of fast MBN */
\FOR{$z=1,\ldots,Z$}
\STATE $\x_{z,i} \leftarrow \y_{i},\mbox{ }\forall i = 1,\ldots,n$\\
\STATE $m\leftarrow 2$
\STATE Randomly generate $\delta$ from the range $[0.05,0.95]$\\
\WHILE{$k_m\ge k_o$}
\FOR{$v=1,\ldots,V$}
\STATE Calculate pairwise similarity matrix $\mathbf{B} = \mathbf{X}_z^T\mathbf{X}_z$ where $\mathbf{X}_z = [\mathbf{x}_{z,1},\ldots,\mathbf{x}_{z,n}]$\\
\STATE Randomly select $k_m$ columns of $\mathbf{B}$ to form a new matrix $\mathbf{B}'$, which is the similarity scores between the input data and the centroids of the $v$-th clustering at the $m$-th layer\\
    \FOR{$i=1,\ldots,n$}
        \STATE Find the largest element of the $i$th row of $\mathbf{B}$, supposed to be the $j$th element\\
        \STATE Derive a one-hot code $\s_{i,v}= [s_{i,v,1},\ldots,s_{i,v,k_m}]^T$ where
        \begin{equation}
          s_{i,v,t}=\left\{
          \begin{array}{ll}
            1, &\mbox{if } t = j\\
            0, &\mbox{otherwise} \\
          \end{array},\mbox{ } \forall t = 1,\ldots,k_m
          \right.\nonumber
        \end{equation}\\ 
    \ENDFOR
\ENDFOR
    \STATE $\x_{z,i}\leftarrow [\s_{i,1}^T,\ldots,\s_{i,k_m}^T]^T,\mbox{ } \forall i = 1,\ldots,n$
\STATE $k_{m+1}\leftarrow \delta k_m$\\
\STATE $m\leftarrow m+1$\\
\ENDWHILE
\STATE $\bar{\y}_{z,i}\leftarrow \bar{\x}_{z,i},\mbox{ } \forall i = 1,\ldots,n,\mbox{ }\forall z= 1,\ldots,Z$
\ENDFOR
    \STATE $\bar{\y}_{i}\leftarrow [\y^T_{1,i},\y^T_{2,i},\ldots,\y^T_{Z,i}]^T,\mbox{ } \forall i = 1,\ldots,n$
\end{algorithmic}
\end{algorithm}

To reduce the computational complexity of MBN-E, we design a new algorithm fMBN-E in Algorithm \ref{alg:fmbn-e}. Its architecture is shown in Fig. \ref{fig:x}. Specifically, fMBN-E and MBN-E differs in the following two aspects.
\begin{itemize}
  \item \textbf{The first novel aspect:} fMBN-E trains a single bottom layer, instead of training $Z$ independent bottom layers as that in MBN-E.
  \item \textbf{The second novel aspect:} For training each MBN base model, fMBN-E removes the random feature selection step from MBN. This modification makes us able to train the MBN base learners by random resampling of similarity scores, instead of random resampling of data.
\end{itemize}

From the above algorithm, we can easily obtain that:

\begin{thm}\label{thm:3}
  \textit{The computational complexity of fMBN-E is $\O(\alpha kVn)+\O(Zn^2)$.}
\end{thm}
Comparing Theorems \ref{thm:2} and \ref{thm:3}, we see that the computational complexities of the bottom layer and the other layers are reduced by $Z$ and $kV/n$ times respectively. For example, in a typical setting where $k=n/2$, $Z = 40$, and $V=400$, the computational complexity of MBN-E is as high as $ (\O(8000\alpha n^2)+\O(8000n^2))$, while the complexity of fMBN-E is $\O(200\alpha n^2)+\O(40n^2)$ which may be hundreds of times faster than MBN-E. Particularly, because the complexity of the original MBN model is $(\O(\alpha kVn)+\O(kVn))$ \cite{zhang2018multilayer}, we can see that fMBN-E may be even faster than a single MBN described in \cite{zhang2018multilayer} since that $V$ is larger than $Z$ in practice.

\subsection{Analysis}\label{discussion}

Here we explain theoretically how the two novel aspects of fMBN-E reduce the computational complexity of MBN-E without suffering significant performance degradation.

\subsubsection{On the first novel aspect of fMBN-E}
\begin{figure}[t]
\centering
\resizebox{5cm}{!}{\includegraphics*{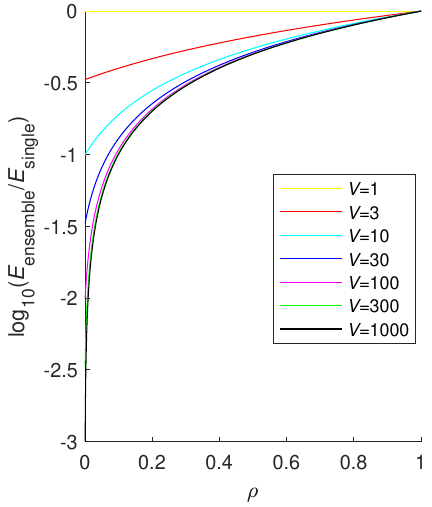}}
\caption{Relationship between the estimation error $\mathbb{E}_{\mathrm{ensemble}}/\mathbb{E}_{\mathrm{single}}$, correlation coefficient $\rho$, and number of $k$-centroids clusterings per layer $V$.}
 \label{fig:jfoa}
\end{figure}

Based on Theorem \ref{thm:4}, we can draw the connections between $\mathbb{E}_{\mathrm{ensemble}}/\mathbb{E}_{\mathrm{single}}$, $\rho$, and $V$ in Fig. \ref{fig:jfoa}, and further derive the following corollary from \eqref{eq:important}.

\begin{cor}\label{cor:0}
The estimation errors of the bottom layers of fMBN-E $\mathbb{E}_{\mathrm{fMBN-E}}$ and MBN $\mathbb{E}_{\mathrm{MBN-E}}$ have the following connection:
   \begin{equation}
\frac{\mathbb{E}_{\mathrm{fMBN-E}}}{\mathbb{E}_{\mathrm{MBN-E}}} = \frac{\left(\frac{1}{V}+\left(1-\frac{1}{V}\right)\rho\right)\mathbb{E}_{\mathrm{single}}}{\left(\frac{1}{ZV}+\left(1-\frac{1}{ZV}\right)\rho\right)\mathbb{E}_{\mathrm{single}}}=\frac{Z+(ZV-Z)\rho}{1+(ZV-1)\rho}
 \end{equation}
\end{cor}

From Corollary \ref{cor:0}, we can further derive the following corollary:

\begin{cor}\label{cor:1}
When $V$ is large enough, the estimation error of the bottom layer of fMBN-E is similar to that of $Z$ independent bottom layers of MBN-E:
   \begin{equation}
{\mathbb{E}_{\mathrm{fMBN-E}}}\approx {\mathbb{E}_{\mathrm{MBN-E}}}
 \end{equation}
\end{cor}

\begin{proof}
According to Corollary \ref{cor:0}, we see that, when $V$ and $Z$ are both large enough, ${\mathbb{E}_{\mathrm{fMBN-E}}}/{\mathbb{E}_{\mathrm{MBN-E}}}$ is determined by $\rho$. For the first case when $\rho\rightarrow 0$, ${\mathbb{E}_{\mathrm{fMBN-E}}}\approx Z {\mathbb{E}_{\mathrm{MBN-E}}} $; for the second case when $\rho\gg 0$, ${\mathbb{E}_{\mathrm{fMBN-E}}}\approx {\mathbb{E}_{\mathrm{MBN-E}}} $. In the following, we show that the second case is true.

It is easy to know that enlarging $k$ reduces $\mathbb{E}_{\mathrm{single}}$. From \eqref{eq:rho}, we also observe that, when $k$ is enlarged, $\rho$ is enlarged as well. According to Theorem \ref{thm:4}, for the bottom layer of MBN, empirically, setting $k$ to a proper number balances $\mathbb{E}_{\text{single}}$ and $\rho$, which produces the minimum $\mathbb{E}_{\text{ensemble}}$. Here we take the common setting $k=n/2$ and $a=0.5$ as an example. In this setting, we may have $\rho \approx 0.0625$, which supports that ${\mathbb{E}_{\mathrm{fMBN-E}}}\approx {\mathbb{E}_{\mathrm{MBN-E}}} $. Corollary \ref{cor:1} is proved.
\end{proof}

Corollary \ref{cor:1} motivates us to train a single bottom layer as fMBN-E, instead of training $Z$ independent bottom layers as MBN-E.

\subsubsection{On the second novel aspect of fMBN-E}

This subsection explains why fMBN-E is able to discard the random feature selection step of MBN when training the upper layers.

\begin{cor}\label{cor:3}
The random feature selection step has limited effect on the upper layers of the MBN base models of fMBN-E.
\end{cor}

\begin{proof}
For the upper layers of fMBN-E, the parameter $k$ is usually far smaller than $n$, e.g. $k=n/2^3$ at the third layer from bottom-up. According to \eqref{eq:rho} if we remove the random feature selection step by setting $a=1$, we may have $\rho \approx 1/2^6$ . From Fig. \ref{fig:jfoa}, we see that $\mathbb{E}_{\text{ensemble}}$ is far smaller than $\mathbb{E}_{\text{single}}$ when $\rho \approx 1/2^6$. Therefore, we do not need the random feature selection step to further pursue a marginal reduction of $\mathbb{E}_{\text{ensemble}}$.
\end{proof}
Corollary \ref{cor:3} motivates us to remove the random feature selection step at the upper layers of fMBN-E, which provides the opportunity to reduce the computational complexity significantly.

Following a similar explanation with the proof of Corollary \ref{cor:3}, we can obtain:
\begin{cor}\label{cor:2}
The random feature selection step reduces the estimation error of the bottom layer of fMBN-E significantly.
\end{cor}
Corollary \ref{cor:2} motivates us to retain the random feature selection step at the bottom layer of fMBN-E.
%
%
%
%
%
%
%



\section{Unsupervised Network Structure Selection}\label{mbn-s}

In this section, we first present an unsupervised ensemble selection framework for MBN-E in Section \ref{subsec:framework}, and then present MBN-SO and MBN-SD in Sections \ref{subsec:MBN-SO} and \ref{subsec:MBN-SD} respectively.

\subsection{Framework}\label{subsec:framework}

 \begin{algorithm}[t]
    \caption{Unsupervised ensemble selection for MBN-E.}
    \begin{algorithmic}[1]\label{alg:so}
\REQUIRE
Sparse output of MBN-E $\{\bby_i \}_{i=1}^{n}$ and its low-dimensional representation $\{\bbu_i\}_{i=1}^n$; \\
Sparse outputs of the MBN base models $\{\{\y_{z,i} \}_{i=1}^{n}\}_{z=1}^Z$ and their low-dimensional representations $\{\{\u_{z,i}\}_{i=1}^n\}_{z=1}^Z$;\\
Number of selected base models $B$\\
Number of classes $c$ (optional). \\
\ENSURE $\{\bar{\bar{\y}}_{i}\}_{i=1}^n$, $\{\bar{\bar{\u}}_{i}\}_{i=1}^n$.
\IF{$c$ is given}
\STATE $\{l_i\}_{i=1}^n\leftarrow\mathrm{clustering}(\{\bbu_i \}_{i=1}^{n},c)$\\
\FOR{$z=1$ to $Z$}
\STATE $\omega_z \leftarrow f_{\textrm{MBN-SO}}(\{l_i\}_{i=1}^n, \{\u_{z,i} \}_{i=1}^{n})$ \\$\quad$(or $\omega_z \leftarrow f_{\textrm{MBN-SO}}(\{l_i\}_{i=1}^n, \{\y_{z,i}\}_{i=1}^{n})$)\\
\ENDFOR
\ELSE
\FOR{$z=1$ to $Z$}
\STATE $\omega_z \leftarrow f_{\textrm{MBN-SD}}(\{\bby_i \}_{i=1}^{n}, \{\y_{z,i} \}_{i=1}^{n})$ \\$\quad$(or $\omega_z \leftarrow f_{\textrm{MBN-SD}}(\{\bbu_i \}_{i=1}^{n}, \{\u_{z,i} \}_{i=1}^{n})$)\\
\ENDFOR
\ENDIF
\STATE Pick $B$ sparse representations that correspond to the $B$ largest weights of $\{\omega_z\}_{z=1}^Z$, supposed to be $\{\{\x_{b,i} \}_{i=1}^{n}\}_{b=1}^B$ without loss of generality
\STATE $\bar{\bar{\x}}_{i} \leftarrow [\x_{1,i}^T,\ldots,\x_{B,i}^T]^T, \mbox{ } \forall i = 1,\ldots,n$
\STATE $\{\bar{\bar{\y}}_{i}\}_{i=1}^n \leftarrow \mathrm{PCA}(\{\bar{\bar{\x}}_{i}\}_{i=1}^n)$
\end{algorithmic}
\end{algorithm}

Algorithm \ref{alg:so} presents the unsupervised ensemble selection framework for MBN-E. If the number of classes $c$ is given, it adopts MBN-SO to select $B$ effective MBN base models. Specifically, it first conducts clustering on $\{\bby_i\}_{i=1}^n$, which generates a set of predicted labels $\{l_i\}_{i=1}^n$. Then, it calculates a weight $\omega_z$ for the $z$-th MBN base model by an optimization-like criterion $f_{\textrm{MBN-SO}}(\{l_i\}_{i=1}^n, \{\y_{z,i} \}_{i=1}^{n})$. The larger the weight $\omega_z$ is, the more important the corresponding MBN base model is.

 If $c$ is not given, it adopts MBN-SD to select the base models. Specifically, it first calculates the weight $\omega_z$ by evaluating the difference between the distributions $\{\bbx_i \}_{i=1}^{n} $ and $\{\x_{z,i} \}_{i=1}^{n}$ directly via a distribution divergence criterion $f_{\textrm{MBN-SD}}(\cdot)$. After obtaining $\{\omega_z\}_{z=1}^Z$, it concatenates the sparse output of the $B$ ($B\ll Z$) MBN base models whose weights are the $B$ largest ones among $\{\omega_z\}_{z=1}^Z$ into a new sparse representation of data $\{\bar{\bar{\x}}_{i}\}_{i=1}^n$.

Note that there are a vast number of ensemble selection algorithms manipulating on $\{\omega_z\}_{z=1}^Z$. Because this is not the focus of this paper, here we prefer the simple yet effective one.
%
%

%
%
%

\subsection{MBN-SO: Ensemble selection with optimization-like criteria}\label{subsec:MBN-SO}

MBN-SO follows the comparison conclusion on the optimization-like criteria \cite{vendramin2010relative}, and picks four best criteria, which are the silhouette width criterion (SWC), point-biserial (PB), PBM, and variance ratio criterion (VRC), respectively. Because they are defined in Euclidian spaces, MBN-SO takes the low-dimensional representations $\{\y_{z,i}\}_{z=1}^Z$ of the MBN base models for evaluation. Due to the length limitation of the paper, we present the four criteria in Appendix C of the Supplementary Material.



\subsection{MBN-SD: Ensemble selection with distribution divergence criteria}\label{subsec:MBN-SD}

MBN-SD adopts MMD, which is a common distribution divergence criterion in unsupervised domain adaptation, to evaluate the distribution divergence between the outputs of MBN-E and its MBN base models. See Appendix C of the Supplementary Material for the detailed derivation of MMD.

Note that, we have studied many probability distribution divergence criteria in literature, including the Kullback-Leibler divergence, total variance distance, L2-norm distance, Hellinger distance, Wasserstein distance, Bhattacharyya distance, etc. Unfortunately, they do not work for MBN-SD. However, it does not mean that MMD is the only choice, which needs further investigation in the future.

\section{Experiments}\label{experiment}

In this section, we first compare the proposed methods with a number of representative methods on several benchmark datasets in Section \ref{subsec:main_result}. Then, we demonstrate how fMBN-E accelerates MBN-E without sacrificing accuracy in Section \ref{subsec:fMBN-E}, and compare the ensemble selection criteria in Section \ref{subsec:effect1}. Finally, we present the experimental conclusions of some important aspects in Section  \ref{subsec:effect2}.

\subsection{Datasets}

\begin{table}[t]
\caption{\label{table:data_set_info}{{Description of data sets.} The term ``optimal $\delta$'' denotes where the optimal performance of MBN appears by searching $\delta$ from a range of $(0,1)$.}}
\renewcommand{\arraystretch}{1.5}
\centerline{\scalebox{0.75}{
\begin{tabular}{llllll}
\hline
{Name} & {\# samples} & \# dimensions & \# classes & Attribute & Optimal $\delta$\\
\hline
 Dermatology & 366 & 34 & 6 & Biomedical & $ (0, 0.2)$\\
 New-Thyroid & 255 & 5 & 3 & Biomedical & $ (0, 0.35)$\\
 UMIST & 575 & 1024 & 20 & Faces & $ (0.75, 0.85)$\\
 Extended-Yale B & 2414   & 32256  & 38 & Faces & $ (0.6, 0.75)$\\
 COIL20 & 1440   & 4096  & 20 & Images& $ (0.8, 0.9)$\\
 COIL100 & 7200   & 1024  & 100 & Images& $ (0.8, 0.9)$\\
 20-Newsgroups & 18846   & 26214  &20 & Text& $ (0.4, 0.5)$\\
 MNIST & 70000  & 768 & 10 & Images& $(0.35, 0.75)$ \\
\hline
\end{tabular}
}}
\end{table}

We selected 8 benchmark datasets as summarized in Table \ref{table:data_set_info}. For Extended-Yale B, because the luminance of the images dominates the similarity measurement instead of the faces themselves, we preprocessed Extended-Yale B by the dense scale invariant feature transform as in \cite{maggu2020deeply}. For 20-Newsgroups, we extracted the term frequency-inverse document frequency (TF-IDF) text feature. PCA preprocessing was applied to the image datasets, which reduced the original features to 100 dimensions. Cosine similarity measurement was used to measure the similarity between the documents of 20-Newsgroups. All other datasets used Euclidean distance as the similarity measurement. Clustering accuracy (ACC) was used as the evaluation metric.

From the table, we see that the operating range of the optimal $\delta$ of MBN appears at dramatically different positions, which are sufficient to demonstrate how the proposed methods address the network structure selection problem, as well as how the proposed methods behave when comparing with the state-of-the-art referenced methods.

\subsection{Parameter settings}

\begin{table*}[t]
  \centering
  \caption{{ACC comparison between the proposed methods and the state-of-the-art referenced methods.} The results of the referenced methods on the datasets marked with ``$*$'' are copied from their original publications or the ``papers with code'' website. The number in bold denotes the best performance.}
    \scalebox{0.93}{\begin{tabular}{lllll}
    \toprule[1pt]
          & Dermatology & New-Thyroid & UMIST* & Extended-Yale B* \\
    \hline
    kmeans & 0.261  & 0.860  & 0.408  & 0.311  \bigstrut[t]\\
    Rank1 & 0.313 (DREC \cite{zhou2019ensemble}) & 0.863 (Borda \cite{sevillano2007bordaconsensus}) & \textbf{0.769 (DASC \cite{zhou2018deep})} & \textbf{0.992 (DMSC \cite{abavisani2018deep})} \\
    Rank2 & 0.307 (LinkClueE \cite{iam2011link}) & 0.859 (LinkClueE \cite{iam2011link}) & 0.750 (DSC-Net-L2 \cite{ji2017deep}) & 0.973  (DSC-Net-L2 \cite{ji2017deep}) \\
    Rank3 & 0.306 (HGPA \cite{strehl2003cluster}) & 0.853 (ECPCS\_MC \cite{huang2018enhanced}) & 0.732 (J-DSSC \cite{lim2020doubly})) & 0.924 (J-DSSC \cite{lim2020doubly})) \\
    Rank4 & 0.299 (CSPA \cite{strehl2003cluster}) & 0.851 (MCLA \cite{strehl2003cluster}) & 0.728 (DSC-Net-L1 \cite{ji2017deep}) & 0.917 (A-DSSC \cite{lim2020doubly}) \\
    Rank5 & 0.297 (ECPCS\_HC \cite{huang2018enhanced}) & 0.845 (Vote \cite{dimitriadou2002combination}) & 0.725 (A-DSSC \cite{lim2020doubly})) & 0.776 (SSC-OMP \cite{you2016scalable}) \\
    MBN (default) & 0.855  & 0.881  & 0.544  & 0.934  \\
    MBN-E & 0.866  & 0.860  & 0.670  & 0.973  \\
    MBN-SO (VRC) & 0.714  & 0.771  & \textbf{0.767 } & 0.941  \\
    MBN-SD & \textbf{0.947 } & \textbf{0.941 } & 0.547  & 0.909  \bigstrut[b]\\
    \hline
    \rowcolor[rgb]{ .851,  .851,  .851} MBN$^\dagger$  & 0.971  & 0.964  & 0.770  & 0.969  \bigstrut\\
   \toprule[1pt]
    \specialrule{0em}{1pt}{4pt}
    \toprule[1pt]
          & COIL20* & COIL100* & 20-Newsgroups & MNIST* \\
    \hline
    kmeans & 0.679  & 0.511 & 0.416 & 0.527 \bigstrut[t]\\
    Rank1 & \textbf{1.000 (JULE \cite{yang2016joint})} & \textbf{0.911 (JULE \cite{yang2016joint})} & 0.600 (LTM \cite{cai2009probabilistic}) & \textbf{0.979 (N2D \cite{mcconville2021n2d})} \\
    Rank2 & 0.858 (AGDL \cite{zhang2012graph}) & 0.824 (A-DSSC \cite{lim2020doubly}) & 0.523 (DFPA \cite{henao2015deep}) & 0.969 (DDC-DA \cite{ren2020deep}) \\
    Rank3 & 0.858 (GDL \cite{zhang2012graph}) & 0.796 (J-DSSC \cite{lim2020doubly})) & 0.490 (LDA \cite{blei2003latent}) & 0.965 (PSSC \cite{villar2020scattering}) \\
    Rank4 & 0.793 (DBC \cite{li2018discriminatively}) & 0.775 (DBC \cite{li2018discriminatively}) & 0.447 (AnchorFree \cite{fu2018anchor}) & 0.964 (GDL \cite{zhang2012graph}) \\
    Rank5 & N/A   & 0.731 (GDL \cite{zhang2012graph}) & 0.435 (LapPLSI \cite{cai2008modeling}) & 0.939 (SR-K-means \cite{jabi2019deep}) \\
    MBN (default) & 0.795  & 0.683  & 0.623  & 0.964  \\
    MBN-E & 0.929  & 0.832  & 0.584  & 0.964  \\
    MBN-SO (VRC) & \textbf{0.995 } & \textbf{0.908 } & \textbf{0.623 } & 0.964  \\
    MBN-SD & 0.973  & 0.803  & 0.611  & 0.963  \bigstrut[b]\\
    \hline
    \rowcolor[rgb]{ .851,  .851,  .851} MBN$^\dagger$  & 0.994  & 0.901  & 0.623  & 0.965  \bigstrut\\
    \toprule[1pt]
    \end{tabular}
    }%
  \label{tab:addlabel_main}
\end{table*}%

The parameter settings of MBN and the proposed methods are summarized as follows:
\begin{itemize}
  \item \textbf{MBN (default) \cite{zhang2018multilayer}:} We used its default setting as in \cite{zhang2018multilayer}. 
  \item \textbf{MBN-E:} It used 40 MBN base models. The base models of MBN-E used the same parameter setting as MBN except that $\delta$ was randomly selected from $[0.05,0.95]$.
  \item \textbf{fMBN-E:} It is the fast version of MBN-E without performance degradation. It discards the random feature selection step in the upper layers of the MBN base models.
  \item \textbf{fMBN-Ev2: } It is a \textit{variant of fMBN-E} that discards the random feature selection step at the bottom layer, and uses the random resampling of similarity scores instead of the random data resampling to train the bottom layer as its upper layers. It accelerates the training time of the bottom layer of fMBN-E, with a risk of performance degradation.
  \item \textbf{MBN-SO:} The number of selected base models $B$ was set to 3. The MBN-SO with the four optimization-like criteria are denoted as ``MBN-SO (SWC)'', ``MBN-SO (PB)'', ``MBN-SO (PBM)'', and ``MBN-SO (VRC)'', respectively.
  \item \textbf{MBN-SD:} The parameter $B$ was set to $10$.
\end{itemize}
Agglomerative hierarchical clustering (AHC) was used for partitioning data into clusters. Although the MMD criterion in MBN-SD is designed to handle the case where the number of classes is unknown, we still give AHC the number of classes during the clustering stage, for a comparable study on how the distribution divergence criterion differs from the optimization-like criteria in MBN-SO. All reported results are average ones over 5 independent runs. The time efficiency was evaluated on an Intel(R) Xeon(R) Platinum 8160 CPU server with 512 GB memory, where the CPU has 48 physical cores. All experiments were run with 48 parallel workers of MATLAB. The source code is available at \url{http://www.xiaolei-zhang.net/mbn-e.htm}.

\subsection{Comparison methods}

The comparison strategy is described as follows. For the image datasets, we copied the ranking lists of the image clustering methods from {\url{https://paperswithcode.com/}}, which reflects the state-of-the-art performance on the datasets. {Note that because self-supervised deep learning based methods explore strong handcrafted features from augmented data \cite{chen2020simple}, we omit them from the experiments to maintain the fairness of the comparison.} For the small-scale Dermatology and New-Thyroid datasets that deep learning methods usually do not handle with, we compared with 12 representative clustering ensemble methods, see Supplementary Material for the referenced methods. All these clustering ensemble methods are meta-clustering functions, which can be used jointly with any base clusterings, such as k-means or spectral clustering. Here we took 40 k-means clusterings as the base clusterings for each meta-clustering function. Like many clustering ensemble methods, e.g. \cite{fred2005combining}, we selected the number of clusters of each  k-means base clustering randomly from a range of $[2c, 10c]$. For the 20-Newsgroups text corpus, we compared with 9 text clustering methods, see \cite{wang2021deep} for the referenced methods. Besides, k-means clustering are also provided as a baseline. Because k-means clustering suffers from bad local minima, we ran k-means clustering on each dataset for 100 times, and pick one that has the minimum objective value. All reported results are average ones over 5 independent runs.

\begin{table*}[t]
  \centering
  \caption{ACC comparison between MBN-E, fMBN-E, and fMBN-Ev2. }
  \scalebox{0.9}{
    \begin{tabular}{lllllllll}
    \hline
          & Dermatology & New-Thyroid & UMIST & Extended-Yale B & COIL20 & COIL100 & 20-Newsgroups & MNIST \bigstrut\\
    \hline
    MBN-E & \textbf{0.866 } & 0.860  & \textbf{0.670 } & \textbf{0.973 } & 0.929  & \textbf{0.832 } & 0.584  & \textbf{0.964 } \bigstrut[t]\\
    fMBN-E & \textbf{0.868 } & \textbf{0.907 } & 0.659  & 0.964  & \textbf{0.938 } & \textbf{0.837 } & 0.582  & \textbf{0.964 } \\
    fMBN-Ev2 & 0.528  & 0.576  & 0.653  & 0.896  & 0.902  & 0.828  & \textbf{0.595 } & 0.963  \bigstrut[b]\\
    \hline
    \end{tabular}%
    }
  \label{tab:afa1}%
\end{table*}%

\begin{table*}[t]
  \centering
  \caption{Running time (in seconds) of the bottom layers of MBN-E, fMBN-E, and fMBN-Ev2. }
   \scalebox{0.9}{
    \begin{tabular}{lllllllll}
    \hline
          & Dermatology & New-Thyroid & UMIST & Extended-Yale B & COIL20 & COIL100 & 20-Newsgroups & MNIST \bigstrut\\
    \hline
MBN-E & 225.08  & 14.96  & 118.00  & 2190.72  & 834.64  & 22148.48  & 59997.16  & 979832.20  \bigstrut[t]\\
    fMBN-E & 0.63  & 0.36  & 3.44  & 70.96  & 24.99  & 679.75  & 1356.35  & 5525.12  \\
    fMBN-Ev2 & 0.84  & 0.74  & 0.82  & 2.74  & 1.17  & 20.58  & 278.06  & 1216.84  \bigstrut[b]\\
    \hline
    \end{tabular}%
    }
  \label{tab:afa2}%
\end{table*}%

\begin{table*}[t]
  \centering
  \caption{Running time (in seconds) of the upper layers of MBN-E, fMBN-E, and fMBN-Ev2.}
   \scalebox{0.9}{
    \begin{tabular}{lllllllll}
    \hline
          & Dermatology & New-Thyroid & UMIST & Extended-Yale B & COIL20 & COIL100 & 20-Newsgroups & MNIST \bigstrut\\
    \hline
MBN-E & 293.85  & 165.15  & 508.75  & 1829.94  & 1413.17  & 5617.11  & 26002.17  & 63939.58  \bigstrut[t]\\
    fMBN-E & 3.02  & 1.63  & 3.38  & 31.85  & 20.05  & 206.46  & 2085.35  & 9108.11  \\
    fMBN-Ev2 & 1.95  & 1.34  & 2.37  & 21.52  & 10.17  & 103.35  & 1141.76  & 8638.58  \bigstrut[b]\\
    \hline
    \end{tabular}%
    }
  \label{tab:afa3}%
\end{table*}%

\subsection{General results}\label{subsec:main_result}

Table \ref{tab:addlabel_main} lists the results of the aforementioned comparison methods and the proposed methods. Because it is too lengthy to list all results, here we only list the results of the top 5 referenced methods; for the proposed MBN-SO variants, we only provide ``MBN-SO (VRC)'' as a representative. See Supplementary Material for the results of the other three variants of MBN-SO. We also list the performance of the MBN with the optimal $\delta$, denoted as MBN$^\dagger$. Note that because it is unlikely to select the optimal $\delta$ manually in real-world applications, MBN$^\dagger$ only provides an upperbound of the proposed methods.

From the table, we see that the proposed methods outperform ``MBN (default)'' in general, as what we have targeted to in this paper. Specifically, MBN-E outperforms ``MBN (default)'' on UMIST, Extended Yale B, COIL20, and COIL100 significantly where the optimal operating range of $\delta$ of MBN is far from the default value 0.5. It is also comparable to ``MBN (default)'' on Dermatology and New-Thyroid. As for MNIST and 20-Newsgroups, even if the default $\delta$ happens to be in the optimal operating range, MBN-E can still be competitive to ``MBN (default)'' if the optimal range is wide enough, such as that on MNIST. MBN-SO further improves the performance of MBN-E, and outperforms ``MBN (default)'' significantly on most datasets, except the small-scale Dermatology and New-Thyroid. Finally, MBN-SD outperforms ``MBN (default)'' on Dermatology and New-Thyroid, COIL20, and COIL100 significantly, and is comparable to the latter in the remaining four datasets.

\begin{figure*}[t]
\centering
\resizebox{14cm}{!}{\includegraphics*{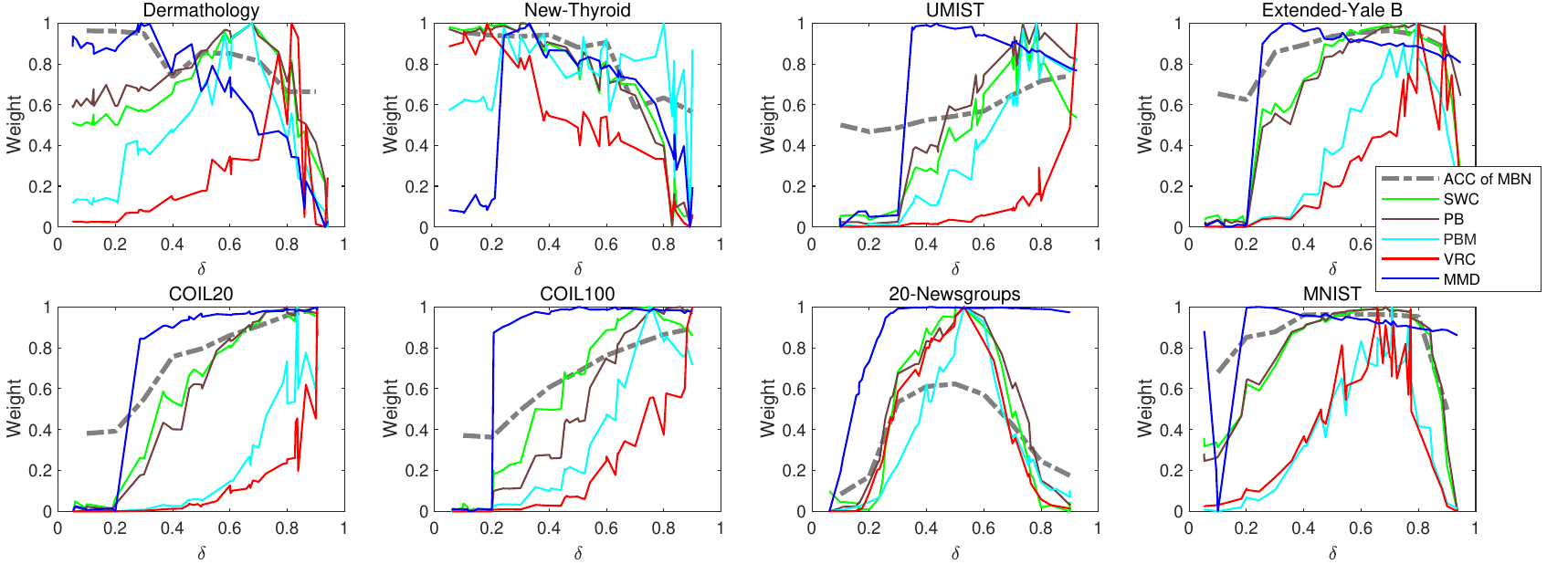}}
\caption{{Weights of the MBN base models produced by different ensemble selection criteria, where SWC, PB, PBM and VRC are optimization-like criteria for MBN-SO, and MMD is a distribution divergence criterion for MBN-SD. The dotted lines in grey color are the accuracies of MBN with respect to $\delta$, which are references for evaluating the effectiveness of the weights.}}
 \label{fig:scores}
\end{figure*}

The proposed MBN-SO also approaches to the top performance of the referenced methods on most datasets. Although it behaves worse than DMSC on Extended Yale B, it still ranks among the top 5 comparison methods. Here we need to emphasize one merit of MBN-SO: it is implemented in a simple mathematical form and behaves robustly across datasets without carefully selected architectures or hyperparameters, which fascinates its practical use. Note that it is interesting to observe that the clustering ensemble methods do not show significant performance improvement over k-means on the small scale Dermatology and New-Thyroid data. Note also that {the performance of text clustering is strongly related to text features. If bag-of-words is used instead of TF-IDF, then the performance of all referenced methods on 20-Newsgroups degrades significantly. To improve the performance on text clustering, new text features that incorporate context information of words may be helpful.} 

Focusing on our three algorithms, we see that MBN-SO is at least comparable to MBN-E and MBN-SD on most of the challenging data, except the two small-scale data where a shallow network of MBN is able to produce a highly accurate result. Comparing MBN-E and MBN-SD, we see that MBN-SD outperforms MBN-E on the two small-scale data, COIL20 and 20-Newsgroups, and is inferior to the latter on UMIST, Extended Yale B, and COIL100. Although the result of MBN-SD is not very impressive, it introduces a new class of ensemble selection criteria---distribution divergence criteria--- into clustering ensemble, which may motivate new criteria beyond MMD for further improving the performance of MBN-SD.

\subsection{Comparison between MBN-E and fMBN-E}\label{subsec:fMBN-E}

Table \ref{tab:afa1} lists the clustering accuracies of MBN-E, fMBN-E, and fMBN-Ev2. From the table, we see that MBN-E and fMBN-E achieve similar performance. This phenomenon supports the correctness of Corollaries \ref{cor:1} and \ref{cor:3}. Moreover, fMBN-E behaves better than fMBN-Ev2, particularly on Dermatology, New-Thyroid, and Extended Yale-B, which supports the correctness of Corollary \ref{cor:2}.

Tables \ref{tab:afa2} and \ref{tab:afa3} summarize the running time of the comparison methods. From the tables, we see that fMBN-E is dozens of times faster than MBN-E on training the bottom layers. Moreover, fMBN-E and fMBN-Ev2 are even hundreds of times faster than MBN-E on training the upper layers. The phenomenon supports the theoretical analysis of Theorem \ref{thm:3}.


\subsection{Comparison between different ensemble selection criteria for MBN-SO and MBN-SD}\label{subsec:effect1}
To study how different ensemble selection criteria affect the weights of the MBN base models, we compared the weights with the clustering accuracy of the MBN base models in a single run in Fig. \ref{fig:scores}. From the figure, we see that the weights produced by all ensemble selection criteria can cleverly reflect the quality of the base models on most datasets except Dermatology. Particularly, the weights produced by ``VRC'' seem to be the most accurate among the ensemble selection criteria. Although the weights produced by ``MMD'', which is a distribution divergence criterion, seem not as accurate as the optimization-like criteria, if we pick a number of MBN base models, then the optimal MBN base models may be selected as well.

\subsection{Discussions}\label{subsec:effect2}

This subsection reports the main conclusions of some important aspects, leaving the detailed description of the experiments in Appendix D of the Supplementary Material.

\subsubsection{Effect of number of selected base models on MBN-SO and MBN-SD}

To study how the number of MBN base models affect the performance of MBN-SO and MBN-SD, we tuned the hyperparameter $B$ from 1 to 10.
 We find that, for MBN-SO, we can set the hyperparameter $B$ to a small number for saving the computing resource; however, for MBN-SD, we should set $B$ to a large number in order to achieve the optimal performance.

\subsubsection{Effect of the referenced labels on MBN-SO}

MBN-SO need referenced labels to calculate the weights of the MBN base models, where we adopt the predicted labels from MBN-E as the reference. After studying different generation methods of referenced labels, including (i) randomly generated labels, (ii) predicted labels from ``MBN (default)'', (iii) predicted labels from MBN-E, and (iv) ground-truth labels, we find that the accuracy of the referenced labels has significant impact on the performance, and that the predicted labels generated from MBN-E yield good performance.

 \subsubsection{On candidate meta-clustering functions of MBN-E}
It is known that combining the base clusterings via a meta-clustering function is important for clustering ensemble technologies. In this paper, we combine the MBN base models by simply concatenating their sparse output without referring to an advanced meta-clustering function. In the Supplementary Material, we have tried 12 representative meta-clustering functions to fuse the output of the MBN base models. Empirical results show that simply concatenating the outputs of the MBN base models yields similar performance to the best meta-clustering functions.

\subsubsection{On candidate ensemble selection methods of MBN-SO}

MBN-SO simply selects the MBN base models with the highest weights. In literature, there are many studies on how to select the base models given the weights, which may lead to higher performance and lower computational power than the proposed method. In the Supplementary Material, we have compared with 8 representative ensemble selection methods as well as their 5 variants. Empirical results show that simply picking the top MBN base models is enough to reach the highest performance, while further exploring the diversity between the base models via complicated ensemble selection algorithms is unnecessary.

%

%
%
%
%

\section{Applications}\label{sec:appl}

In this section, we apply the proposed algorithms to image segmentation and graph data mining.

\begin{figure*}[t]
\centering
\resizebox{13cm}{!}{\includegraphics*{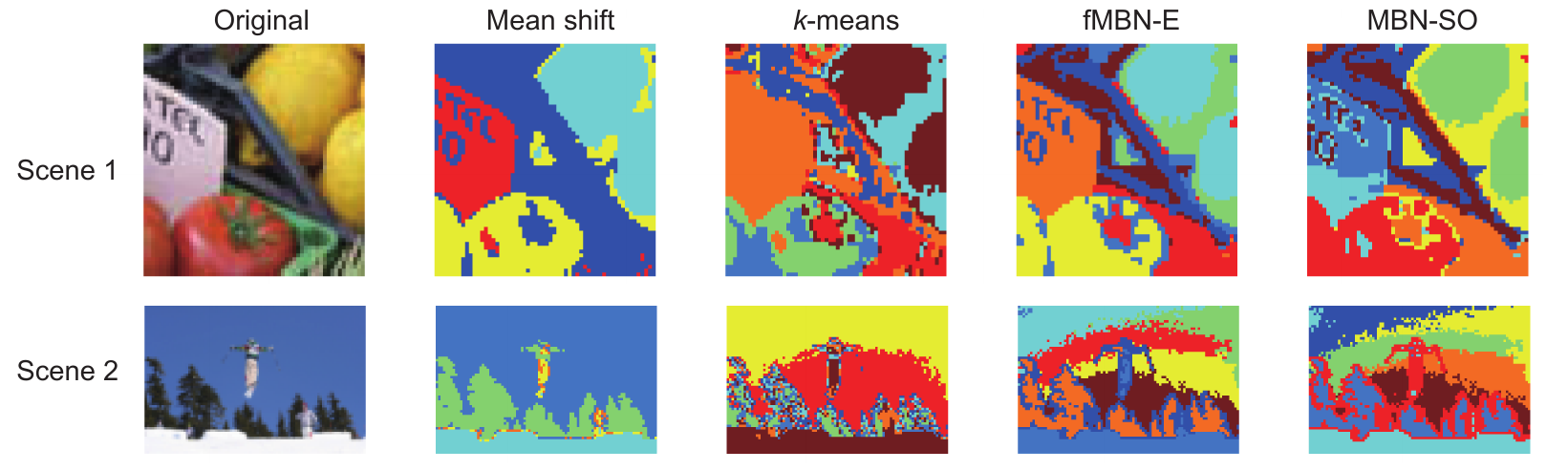}}
\caption{{Results of the image segmentation methods on 2 randomly selected examples from the 2017 Val images of the COCO datasets. }}
 \label{fig:imageseg}
\end{figure*}

\subsection{Application to image segmentation}

Image segmentation partitions an image into multiple image segments, so as to simplify the analysis of the image. It is a process of assigning a label to every pixel of an image such that the pixels with the same label share certain characteristics. It is a core task of image signal processing. It can be either unsupervised or supervised. Unsupervised image segmentation, which is usually used as a preprocessing of supervised segmentation, is formulated as a clustering problem on pixels such that the pixels with similar colors and nearby locations are grouped into the same cluster.

\begin{table}[t]
  \centering
  \caption{Description of the GEMSEC-facebook datasets. }
   \scalebox{0.9}{
    \begin{tabular}{lccc}
    \hline
          & Number of nodes & Density & Transitivity  \bigstrut\\
    \hline
Politicians& 5,908& 0.0024& 0.3011\\
Companies& 14,113& 0.0005& 0.1532\\
Athletes& 13,866& 0.0009& 0.1292\\
News sites& 27,917& 0.0005& 0.1140\\
Public figures& 11,565& 0.0010& 0.1666\\
Artists &50,515& 0.0006& 0.1140\\
Government& 7,057& 0.0036& 0.2238\\
TV shows& 3,892& 0.0023& 0.5906\\
    \hline
    \end{tabular}%
    }
  \label{tab:fjaow}%
\end{table}%

\begin{table*}[t]
  \centering
  \caption{Modularity of the community detection algorithms on the GEMSEC-facebook datasets. The results of the referenced methods are copied from \cite{rozemberczki2019gemsec}.}
   \scalebox{0.9}{
    \begin{tabular}{lccccccccc}
    \hline
& Politicians &Companies& Athletes& News sites &Public figures& Artists &Government &TV shows & \textbf{Ranking}\\\hline
\multirow{2}*{Overlap factorization \cite{ahmed2013distributed}} & 0.810 &0.553 &0.601 &0.471 &0.551 &0.474 &0.608  &0.786& \multirow{2}*{\textbf{4.57}}\\
&\scriptsize{($\pm$0.008)} &\scriptsize{($\pm$0.010)} &\scriptsize{($\pm$0.020)} &\scriptsize{($\pm$0.016)} &\scriptsize{($\pm$0.01)} &\scriptsize{($\pm$0.018)} &\scriptsize{($\pm$0.024)} &\scriptsize{($\pm$0.008)}\\
\multirow{2}*{Walktrap \cite{pons2005computing}} &0.841 &0.639 & 0.670 &0.514 &0.628 &0.554 &0.675 &0.790& \multirow{2}*{\textbf{2.00}}\\
&\scriptsize{($\pm$0.023)} &\scriptsize{($\pm$0.016)} &\scriptsize{($\pm$0.021)} &\scriptsize{($\pm$0.023)} &\scriptsize{($\pm$0.023)} &\scriptsize{($\pm$0.026)} &\scriptsize{($\pm$0.043)} &\scriptsize{($\pm$0.036)}\\
\multirow{2}*{Fast greedy \cite{clauset2004finding}} &0.819 &0.665& 0.605& 0.531 &0.630 &0.464& 0.615& 0.835& \multirow{2}*{\textbf{2.86}}\\
&\scriptsize{($\pm$0.008)} &\scriptsize{($\pm$0.014)} &\scriptsize{($\pm$0.026)} &\scriptsize{($\pm$0.020)} &\scriptsize{($\pm$0.011)} &\scriptsize{($\pm$0.023)} &\scriptsize{($\pm$0.046)} &\scriptsize{($\pm$0.006)}\\
\multirow{2}*{Label propagation \cite{gregory2010finding}} &0.826 &0.647 &0.647 &0.243 &0.612 &0.393& 0.659& 0.839& \multirow{2}*{\textbf{3.29}}\\
&\scriptsize{($\pm$0.009)} &\scriptsize{($\pm$0.075)} &\scriptsize{($\pm$0.094)} &\scriptsize{($\pm$0.159)} &\scriptsize{($\pm$0.027)} &\scriptsize{($\pm$0.018)} &\scriptsize{($\pm$0.041)} &\scriptsize{($\pm$0.004)}\\
\multirow{2}*{fMBN-E} &0.830&	0.549&	0.657&	0.518&	0.580&	0.502&	0.681&	0.809& \multirow{2}*{\textbf{2.29}}\\
&\scriptsize{($\pm$0.004)} &	\scriptsize{($\pm$0.011)} &	\scriptsize{($\pm$0.002)} &	\scriptsize{($\pm$0.014)} &	\scriptsize{($\pm$0.015)} &	\scriptsize{($\pm$0.003)} &	\scriptsize{($\pm$0.009)} &	\scriptsize{($\pm$0.005)} \\
    \hline
    \end{tabular}%
    }
  \label{tab:dadaa}%
\end{table*}

We randomly selected several images from the 2017 Val images of the COCO datasets\footnote{https://cocodataset.org} for evaluation. We reduced the length and width of each image to about 1/7 of their original sizes, and further transformed the color space from RGB to CIELAB. Finally, for each pixel, we concatenated its three-dimensional colors and its two-dimensional coordinates as the feature. We compared with the classic mean-shift clustering and k-means clustering. The bandwidth of mean-shift was set to 0.2. The clustering number of both k-means clustering and the proposed methods was set to 8. We applied k-means clustering to the output of the proposed methods.

Two examples of the comparison results are shown in Fig. \ref{fig:imageseg}, while more examples are listed in Appendix E of the Supplementary Materials. From the figure, we see that the proposed methods not only maintain sufficient details of the images than mean-shift, but also yield smoother and more accurate results than k-means. As for the proposed methods, MBN-SO behaves similarly to fMBN-E.

\subsection{Application to graph data mining}

All of the aforementioned experiments were conducted on the data whose features are given explicitly. However, the data points in many real-world applications do not have explicit features, e.g. graph data where only the connections between the data points are given. Here we give an example on how to apply the proposed methods to graph data.

Community detection is a method for finding groups within complex systems that are represented on a graph. It is a core task of network science, and finds its applications in network security, recommendation systems, etc. As collected in \url{https://snap.stanford.edu/data/}, the data in community detection are various sparse graphs. Here we used the undirected GEMSEC-facebook data in the collection for evaluation.

The statistics of the GEMSEC-facebook data is summarized in Table \ref{tab:fjaow}. For each link between a node $i$ and a node $j$, we set the elements $b_{i,j}$ and $b_{j,i}$ of the graph $\mathbf{B}$ to the weight of the link. Because the pairwise similarity between the nodes has already been given as $\mathbf{B}$, the output of each $k$-centroids clustering at the bottom layer of fMBN-E is simply a random sample of the columns of $\mathbf{B}$. Because the ground-truth number of communities is unknown, we used \textit{modularity} as the evaluation metric as that in \cite{rozemberczki2019gemsec}. Because the modularity can be calculated in an unsupervised manner by comparing $\mathbf{B}$ with the prediction result, we are able to search for the optimal modularity results as \cite{rozemberczki2019gemsec}. Specifically, we set parameter $k_o$ of fMBN-E to $1.5c$ where $c$ was set to 10, 20, 30, and 40 respectively. For each $k_o$, we grouped the nodes to 2 to 50 communities, and picked the optimal result in terms of the modularity. We applied k-means clustering to the output of the proposed methods. Following \cite{rozemberczki2019gemsec}, we reported the average results over 5 independent runs. Table \ref{tab:dadaa} lists the comparison results with four well-known community detection algorithms \cite{ahmed2013distributed,pons2005computing,clauset2004finding,gregory2010finding}. From the average ranking over the 8 community detection tasks, we see that the proposed fMBN-E ranks the second, which is slightly worse than the walktrap algorithm \cite{pons2005computing}. Note that because MBN-SD yields almost identical performance with fMBN-E, we omit its result here.


\section{Conclusions}\label{conclusion}

In this paper, we aim to derive a simple and tuning-free deep clustering tool that is able to yield comparable performance to the state-of-the-art deep clustering methods, for the sake of towards solving the heavy parameter-tuning problem in clustering. To achieve this goal, we propose to automatically determine the network structure of the deep clustering algorithm---MBN---by ensemble learning and selection. The proposed MBN-E simply concatenates the sparse output of a number of MBN base models with different $\delta$ to a meta-representation. The proposed MBN-SO and MBN-SD use the output of MBN-E to select the base models whose output distributions have the highest discriminability, without further exploring the diversity between the base models as conventional ensemble selection methods did. Because training an ensemble of MBN is expensive, we proposed fMBN-E, which first discards the random feature selection step of MBN and then replaces the step of random data resampling by the random resampling of similarity scores. We proved theoretically that this simplification does not degrade the estimation accuracy of MBN-E. Finally, the above methods contribute an efficient off-the-shelf deep clustering tool.

Experimental comparison results on a wide variety of benchmark datasets show that the proposed methods significantly outperform the MBN with the default network structure; fMBN-E is empirically hundreds of times faster than MBN-E without suffering performance degradation; MBN-SO is able to detect the optimal MBN base model, and reaches comparable performance to the state-of-the-art clustering methods; although MBN-SD is less effective than MBN-SO, it is the first work of unsupervised ensemble selection based on the distribution divergence criteria. Further studies also show that the proposed methods reach top performance via only a simple formulation, comparing to as many as 20 candidate meta-clustering functions and clustering ensemble selection functions. At last, we show the advantage of the proposed methods in image segmentation and graph data mining.

\bibliography{zxlrefs,mywork}
\bibliographystyle{IEEEtran}

\end{document}